\pdfoutput=1

\documentclass[11pt]{article}

\usepackage{EMNLP2023}

\usepackage{times}
\usepackage{latexsym}

\usepackage[T1]{fontenc}

\usepackage[utf8]{inputenc}

\usepackage{microtype}

\usepackage{inconsolata}

\usepackage{amssymb}  
\usepackage{pifont}   
\usepackage{wasysym}  
\usepackage{fontawesome}

\newcounter{experimentcounter}

\usepackage{graphicx}
\usepackage{float}
\usepackage{placeins}

\usepackage{cuted}
\usepackage{multirow}
\usepackage{multicol}
\usepackage{booktabs}
\usepackage{listings}
\usepackage{xcolor}
\usepackage{framed}
\usepackage{lipsum}

\definecolor{shadecolor}{RGB}{240,240,240}
\definecolor{codegreen}{rgb}{0,0.6,0}
\definecolor{codegray}{rgb}{0.5,0.5,0.5}
\definecolor{codepurple}{rgb}{0.58,0,0.82}
\definecolor{backcolour}{rgb}{0.95,0.95,0.92}

\lstdefinestyle{mystyle}{
  backgroundcolor=\color{backcolour}, commentstyle=\color{codegreen},
  keywordstyle=\color{magenta},
  numberstyle=\tiny\color{codegray},
  stringstyle=\color{codepurple},
  basicstyle=\ttfamily\footnotesize,
  breakatwhitespace=false,         
  breaklines=true,                 
  captionpos=b,                    
  keepspaces=true,                 
  numbers=left,                    
  numbersep=5pt,                  
  showspaces=false,                
  showstringspaces=false,
  showtabs=false,                  
  tabsize=2
}

\title{RAGLAB: A Modular and Research-Oriented Unified Framework for Retrieval-Augmented Generation}

\author{
\textbf{Xuanwang Zhang}$^{1,2,*}$,
\textbf{Yunze Song}$^{2,*}$,
\textbf{Yidong Wang}$^{3,2}$ ,
\textbf{Shuyun Tang}$^5$ , \\
\textbf{Xinfeng Li}$^4$ ,
\textbf{Zhengran Zeng}$^3$ ,
\textbf{Zhen Wu}$^{1,\dagger}$,
\textbf{Wei Ye}$^3$ ,  
\textbf{Wenyuan Xu}$^4$ ,  \\
\textbf{Yue Zhang}$^6$ ,
\textbf{Xinyu Dai}$^1$ ,
\textbf{Shikun Zhang}$^3$ ,
\textbf{Qingsong Wen}$^2$ \\
$^1$Nanjing University \hfill    $^2$Squirrel AI \hfill    $^3$Peking University , \\
$^4$Zhejiang University \hfill    $^5$Google \hfill    $^6$Westlake University \\
\texttt{\{zxw.ubw, YunzeSong77\}@gmail.com} \\
}

\begin{document}
\maketitle
\footnotetext[1]{$^*$Equal contribution}
\footnotetext[2]{$^\dagger$Corresponding author}
\begin{abstract}
Large Language Models (LLMs) demonstrate human-level capabilities in dialogue, reasoning, and knowledge retention. However, even the most advanced LLMs face challenges such as hallucinations and real-time updating of their knowledge. Current research addresses this bottleneck by equipping LLMs with external knowledge, a technique known as Retrieval Augmented Generation (RAG). However, two key issues constrained the development of RAG. First, there is a growing lack of comprehensive and fair comparisons between novel RAG algorithms. Second, open-source tools such as LlamaIndex and LangChain employ high-level abstractions, which results in a lack of transparency and limits the ability to develop novel algorithms and evaluation metrics. To close this gap, we introduce RAGLAB, a modular and research-oriented open-source library. RAGLAB reproduces 6 existing algorithms and provides a comprehensive ecosystem for investigating RAG algorithms. Leveraging RAGLAB, we conduct a fair comparison of 6 RAG algorithms across 10 benchmarks. With RAGLAB, researchers can efficiently compare the performance of various algorithms and develop novel algorithms. 

\href{https://github.com/fate-ubw/RAGLab}{\faGithub\, https://github.com/fate-ubw/RAGLab}

\end{abstract}

\section{Introduction}

Retrieval augmentation generation(RAG) leverages external knowledge to mitigate hallucination issues, ensure real-time knowledge updates, and protect private data with no parametric knowledge\citep{naive-rag-2017-qa-baseon-wiki,naiverag-2020-1,naiverag-2-Realm}. 
However, researchers face two main barriers to investigating new RAG algorithms. On the one hand, many published works are either not open-source or have difficulty setting up the environment. While open-source works lack modular design, it is hard to develop new algorithms or extend new datasets for evaluation. Researchers have to waste a lot of time developing new algorithms from scratch. On the other hand, a multitude of novel RAG algorithms have merged, including ITER-RETGEN\citep{iter-retgen}, RRR\citep{rrr-rag}, Self-Ask\citep{self-ask}, Active RAG\citep{active-rag}, Self-RAG\citep{selfrag}, etc. However, these RAG algorithms are not well aligned in their fundamental components and evaluation methodologies, making it difficult for researchers to accurately assess their improvements. As a result, the absence of a unified framework makes it difficult for researchers and engineers to select appropriate algorithms for varied contexts, potentially hindering the advancement of the field.

\begin{table*}[ht]
\centering
\caption{Comparison of Different RAG Libraries and Frameworks. Fair Comparison refers to aligning all fundamental components during evaluation, including random seeds, generator, retriever, and instructions. Data Collector refers to the ability to gather or generate training and test data, either by sampling from existing raw datasets or by constructing labeled data using LLMs.}
\small
\renewcommand{\arraystretch}{1.1}
\setlength{\tabcolsep}{4pt}
\begin{tabular}{lccccc}
\hline
\textbf{Library} & \textbf{Fair Comparison*} & \textbf{Data Collector*} & \textbf{Trainer} & \textbf{Auto Evaluation} & \textbf{Modular Design} \\
\hline
Langchain\citep{LangChain2022}  & \ding{55} & \ding{55} & \ding{55} & \ding{55} & \checkmark \\
LlamaIndex\citep{LlamaIndex} & \ding{55} & \ding{55} & \ding{55} & \checkmark  & \checkmark \\
Haystack\citep{haystack}   & \ding{55} & \ding{55} & \ding{55} & \checkmark & \checkmark \\
FastRAG\citep{fastRAG2023}    & \ding{55} & \ding{55} & \ding{55} & \ding{55} & \checkmark \\
RALLE\citep{ralle-framework}       & \ding{55} & \ding{55} & \ding{55} & \ding{55} & \checkmark \\
LocalRQA\citep{localrqa}   & \ding{55} & \checkmark & \checkmark & \checkmark & \ding{55} \\
AutoRAG\citep{AutoRAG2024}    & \ding{55} & \ding{55} & \ding{55} & \checkmark & \checkmark \\
FlashRAG\cite{FlashRAG}   & \ding{55} & \ding{55} & \ding{55} & \checkmark & \checkmark \\
RAGLAB(ours)     & \checkmark & \checkmark & \checkmark & \checkmark & \checkmark \\
\hline
\end{tabular}
\label{tab:1-comparision_diff_framework}
\end{table*}

While various current works are investigating these questions, such as LlamaIndex \citep{LlamaIndex}, LangChain\citep{LangChain2022}, Haystack\citep{haystack}, FastRAG\citep{fastRAG2023}, RALLE \citep{ralle-framework}, LocalRQA\citep{localrqa}, AutoRAG\citep{AutoRAG2024}, and FlashRAG\citep{FlashRAG}.
LlamaIndex, LangChain, and Haystack are excessively encapsulation and lack transparency in internal operational mechanisms. Consequently, even experienced experts abandon tools like LangChain due to the lack of transparency\citep{woolf-abandon-langchain-2023}. FastRAG and RALLE offer light and transparent frameworks that enable users to assemble their own RAG systems using core components. AutoRAG provides comprehensive metrics to assist users in selecting an optimal RAG system for customized data. LocalRAG provides a wide selection of model training algorithms and evaluation methods. However, LocalRAG, FastRAG, AutoRAG, and RALLE do not reproduce published algorithms. Researchers still need to invest time in replicating algorithms using the provided components. FlashRAG addressed this issue by reproducing a substantial number of existing algorithms. However, FlashRAG lacks training functionalities and fails to properly align generators during inference, leading to unfair comparisons among various algorithms. For a more detailed comparison, refer to Table~\ref{tab:1-comparision_diff_framework}.

To close this gap, we present RAGLAB, a researcher-oriented RAG toolkit for a fair comparison of existing RAG algorithms and simplify the process of developing new algorithms.
RAGLAB provides a modular architecture for each component of the RAG system, providing an ideal platform for fair comparison of algorithms. Additionally, RAGLAB designs an interactive mode and user-friendly interface, facilitating both educational purposes and demonstrations.

In this paper, we introduce the RAGLAB framework, giving an overview of core components and system workflows(section~\ref{sec:2-RAGLAB}).
We standardized key experimental variables: generator fine-tuning, instructions, retrieval configurations, knowledge bases, and benchmark. As a result, we present a comprehensive and fair comparison of 6 RAG algorithms across 10 benchmarks(section~\ref{sec:3-Experiment}).

RAGLAB is available on GitHub under the MIT license.

\begin{figure*}[t]
    \centering
    \includegraphics[scale=0.54]{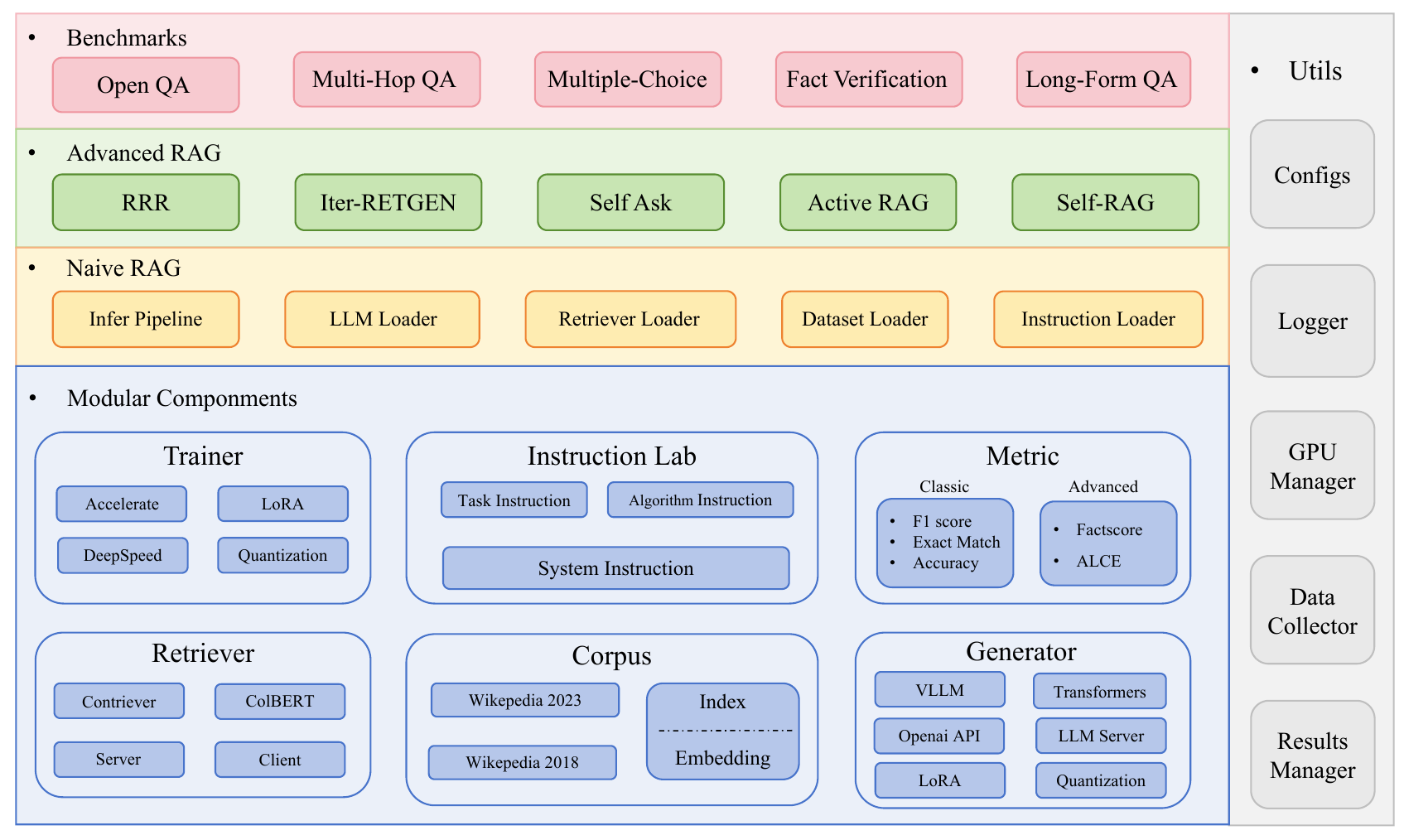}
    \caption{Architecture and Components of the RAGLAB Framework.}
    \label{fig:ragalb-system}
\end{figure*}

\section{RAGLAB}
\label{sec:2-RAGLAB}
The overall architecture of RAGLAB is illustrated in Figure~\ref{fig:ragalb-system}. We first introduce the core classes and concepts, then demonstrate an experimental case using a concise 5-line code snippet.
\subsection{Classes and Concepts}
\subsubsection{Retriever}
RAGLAB integrates two high-performing BERT-based models, Contriever\citep{contriever-2021} and ColBERT\citep{colbertv2-2022}. Furthermore, RAGLAB unifies the query interfaces across different retriever classes, making it possible for users to seamlessly switch between various retrievers. During the evaluation phase, researchers need to benchmark multiple RAG algorithms in parallel. In this context, repeatedly loading and querying retriever models and knowledge databases consumes a amount of time. 

To address this issue, RAGLAB designs a retriever server and client architecture, enabling high-concurrency access to retrievers. Additionally, RAGLAB implements a retrieval caching mechanism. This mechanism stores the results of initial queries and their retrieved knowledge. Consequently, when queried with an identical input, the retriever will return the cached results directly without recomputation. Based on RAGLAB, users only need to load the retriever model and knowledge database once, facilitating retrieval services with latencies of less than 0.1 seconds across multiple parallel evaluation experiments.

\subsubsection{Corpus}

The external knowledge database has a significant impact on the performance of RAG systems. Wikipedia collects all kinds of knowledge and is broadly used in research areas. However, raw web data must undergo complex preprocessing before it can be directly utilized by RAG systems.

RAGLAB provides preprocessed Wikipedia corpora in two versions: the first version is based on the 2018 Wikipedia data open-sourced by the DPR project\citep{DPR-2020-emnlp}; the second version utilizes the 2023 Wikipedia data open-sourced by the FactScore\citep{factscore-2023}.
RAGLAB also pre-built indices and embeddings for both the ColBERT and Contriever models, based on the Wikipedia 2018 and Wikipedia 2023 corpora. Additionally, RAGLAB open-sources all processing scripts, enabling researchers to directly download the preprocessed Wikipedia corpora along with their corresponding indices and embeddings.

\subsubsection{Generator}

The generator is the core component of the RAG system. We integrate Huggingface Transformers\citep{huggingface-transformers-2020} and VLLM\cite{vllm}, thereby enabling RAGLAB to be compatible with a wide range of open-source models while providing stable and efficient inference performance.
RAGLAB also incorporates quantization and Low-Rank Adaptation (LoRA) \citep{lora} features, enabling users to employ models with 70 billion parameters or more as generators, even with limited computational resources.

Furthermore, anticipating the potential need for users to simultaneously load multiple generators within a single RAG algorithm, we develope a GPU management module. This module enables users to precisely allocate multiple generators across specified GPUs through parameter configurations.

 In addition to open-source models, Generator modular includes \verb|OpenaiModel|, supporting closed-source LLMs such as OpenAI. Future developments will extend support to other closed-source LLMs including Claude, Gemini and Azure.

\begin{figure*}
\begin{lstlisting}[language=Python]
from RAGLAB.rag.infer_alg import SelfRag_Reproduction
from utils import get_config()

args = get_config()
query = "What is Henry Feilden's occupation?" # query for interaction mode
Rag = SelfRag_Reproduction(args)

# interact mode
inference_result, generation_track = rag.inference(query, mode = 'interact')
print(inference_result)
# evaluation mode
evaluation_result = rag.inference(mode = 'evaluation')
print(evaluation_result)
\end{lstlisting}
\caption{A script that uses RAGLAB for reproducing Self-RAG  algorithm.}
\label{fig:raglab-code-demo}
\end{figure*}

\subsubsection{Instruction Lab}

Instruction has a significant impact on the quality of output generated by LLMs\citep{prompt-survey}. However, in frameworks such as LlamaIndex and LangChain, many key instructions lack transparency, being encapsulated at lower levels of the architecture. This encapsulation makes it challenging for users to modify. We find that the majority of published works developing their RAG algorithms utilize unaligned instructions, rendering experimental results across different studies incomparable.

To address these issues, RAGLAB designs the Instruction Lab module, which includes three key components: System Instruction, Task Instruction, and Algorithm Instruction. This module allows users to efficiently import and combine desired prompts from 3 instruction pools. Furthermore, users can adjust parameters within the configuration settings, facilitating comparative experiments using different instructions.

\subsubsection{Trainer}

RAGLAB integrates Accelerate\citep{accelerate} and DeepSpeed libraries to provide comprehensive and efficient fine-tuning capabilities. Additionally, the Trainer module supports Low-Rank Adaptation (LoRA) and Quantized LoRA (QLoRA) \citep{qlora-2023} techniques, enabling users to fine-tune models with 70 billion parameters or more with limited computational resources.

We find that recent studies explore a novel method: adding special tokens during the generator training process to enhance performance. To facilitate the reproduction of these published works, the Trainer module supports adding special tokens during the fine-tuning phase.

\subsubsection{Dataset and Metric}

As shown in Table~\ref{tab:benchmark-datasets}, following a comprehensive investigation, RAGLAB collects 10 widely used benchmarks encompassing five distinct tasks.

\begin{table}[ht]
\centering
\caption{Tasks and Datasets.}
\label{tab:benchmark-datasets}
\footnotesize
\renewcommand{\arraystretch}{1.0}  
\begin{tabular}{cl}
\toprule
Task Type & Benchmarks \\
\midrule
\multirow{2}{*}{OpenQA} & PopQA \citep{Popqa} \\
       & TriviaQA \citep{triviaqa-2017} \\
\addlinespace[0.5em]  
\multirow{2}{*}{Multi-HopQA} & HotpotQA \citep{hotpotqa-2018} \\
            & 2WikiMultiHopQA \citep{2WikiMultiHopQA} \\
\addlinespace[0.5em]
\multirow{2}{*}{Multiple-Choice} & ArcChallenge \citep{ARC} \\
                & MMLU \citep{MMLU} \\
\addlinespace[0.5em]
\multirow{3}{*}{Fact Verification} & PubHealth \citep{PubHealth} \\
                  & StrategyQA \citep{strategyqa-2021} \\
                  & FactScore \citep{factscore-2023} \\
\addlinespace[0.5em]
\multirow{2}{*}{Long-Form QA} & ASQA \citep{asqa-2022} \\
             & FactScore \citep{factscore-2023} \\
\bottomrule
\end{tabular}
\end{table}

RAGLab implements a flexible data adaptation mechanism by individually mapping keys for each dataset, addressing the variability in raw data structures across different datasets. This approach enables users to easily extend new datasets by inheriting from existing dataset classes.

RAGLAB provides 3 classic metrics and 2 advanced metrics. Classic metrics include accuracy, exact match, and F1 score. Advanced Metrics include Factscore\citep{factscore-2023} and ALCE\citep{ACLE-2023}.
More specifically, FactScore represents an advanced metric evaluating the factual accuracy of long-form generation, while ALCE serves as a benchmark for assessing the citation accuracy and recall of RAG systems. Additionally, ALCE integrates other metrics, including ROUGE-L, MAUVE, str-em, and str-hit.

\begin{table*}[ht]
\centering
\caption{Performance comparison of various RAG algorithms using  Llama3-8B as base language model.}
\label{tab:exp-1}
\scriptsize
\setlength{\tabcolsep}{3pt}
\renewcommand{\arraystretch}{1.1}
\begin{tabular*}{\textwidth}{@{\extracolsep{\fill}}lcccccccccccccccccc@{}}
\hline
\multirow{2}{*}{Method} & \multicolumn{2}{c}{PopQA} & \multicolumn{2}{c}{TriviaQA} & \multicolumn{2}{c}{HopPopQA} & \multicolumn{2}{c}{WikiMultiHop} & \multicolumn{2}{c}{ARC} & \multicolumn{2}{c}{MMLU} & \multicolumn{2}{c}{PubHealth} & \multicolumn{2}{c}{StrategyQA} & Factscore & ASQA \\
 & ACC & F1 & ACC & F1 & ACC & F1 & ACC & F1 & ACC & F1 & ACC & F1 & ACC & F1 & ACC & F1 & factscore & str-em \\
\hline
Direct & 25.6& 16.3& 68 & 59.8& 20.6& 23.8 & 24.8& 25.7 & 77.4& 57.2& 58& 42.8& 76& 76& 56.4& 56.4& 55.1 & 25.4\\
NaiveRag & 38.8& 22.2& 64.8& 50.7& 28.2& 25.9 & 18& 22.8 & 69.2& 50.9& 50.8& 37.1& 66.2& 66.2& 58.2& 58.2& 83.7 & 23.8\\
RRR& 38.4& 22.9& 58.4& 45.5& 21.6& 20.3 & 21.4& 22.4 & 69.6& 50.8& 50& 38 & 65.4& 65.4& 57.6& 57.6& 84.0 & 23.6\\
ITER-RETGEN& 34.8& 26.5& 65.4& 50.4& 28.8& 26.9 & 19.8& 24.6 & 68.8& 52.4& 52.6& 39.3& 44.2& 44.2& 57 & 57 & 81.4 & 23.1\\
Active Rag & 34.6& 24.2& 64.2& 48.8& 28.8& 26.6 & 16& 22.9 & 66.2& 49.6& 52& 39.2& 41.8& 41.8& 56.8& 56.8& 82.7 & 23.5\\
Self Ask& 11.8& 10.7& 35.6& 27.3& 16.2& 19.6 & 15.2& 19.4 & 55.6& 38.8& 45.4& 32.7& 37.6& 37.07 & 50.4& 50.4& 49.3 & 13.4\\
\hline
\multicolumn{19}{l}{Self-RAG} \\
always retrieval& 38& 12.4& 61.8& 29 & 23& 15.8 & 18.6& 11.6 & 58.6& 41.6& 39.2& 25 & 67.8& 67.8& 48 & 45.8& 70.6 & 32.3 \\
adaptive retrieval & 35.6& 11.2& 56.4& 27.1& 21& 14.4 & 20.4& 13.5 & 58 & 42.6& 39.2& 26.2& 67.8& 67.8& 46.4& 41 & 67.0 & 23.6\\
no retrieval& 14.8& 6.7 & 31.4& 13.2& 11.2& 6.4& 21& 13.33& 58 & 42.6& 39.6& 26.2& 68.4& 68.4& 47.2& 10.2& 29.0 & 7.6\\
\hline
\end{tabular*}
\end{table*}

\subsection{Architecture and Development Guide}

RAGLAB reproduces six published RAG algorithms, encompassing Naive RAG, RRR, ITER-RETGEN, Self-ASK, Active RAG, and Self-RAG.These algorithms share numerous similarities, and each advanced RAG algorithm essentially represents an improvement upon Naive RAG. 

\begin{figure}[ht!]
\begin{lstlisting}[language=Python, basicstyle=\scriptsize]
from raglab.rag.infer_alg import NaiveRag

class NewAlgorithm(NaiveRag):

    def __init__(self, args):
        super().__init__(args)

    def init(self,args):
        # customized  parameters if need
        # customized components if need

    def infer(self, query:str):
        #  pre build components: 
        '''
        self.find_instruction() -> instructions
        self.retrieval.search() -> query
        self.llm.generate() -> output
        ect.
        '''

         # algorithm inference logic 
         
        return output_text
\end{lstlisting}
\caption{Demostriction of developing new RAG algorithms in RAGALB.}
\label{fig:raglab-new algotithm-demo}
\end{figure}

The design philosophy of RAGALB draws inspiration from the HuggingFace Transformer library. Users only need to define their model from the Transformer library, after which they can employ the \verb|generate()| method for inference. RAGALB implements each RAG algorithm as a distinct class. Two critical methods in each algorithm class are \verb|init()| and \verb|infer()|. The \verb|init()| method serves to set parameters and load Generators, while the \verb|infer()| method implements the algorithm's inference process.
Based on this design framework, users can develop new algorithms through a few simple steps, as shown in Figure~\ref{fig:raglab-new algotithm-demo}:
(1) Define a \verb|NewMethod()| class that inherits from \verb|NaiveRAG|.
(2) Add necessary parameters and components for the new algorithm by overriding the \verb|init()| method.
(3) Implement the new algorithm's inference logic by overriding the \verb|infer()| method, utilizing the framework's provided components.

Algorithms inheriting from NaiveRAG can reuse the \verb|inference()| method and all utility functions. Notably, the \verb|inference()| method already provides automatic evaluation and interaction functionalities. This design enables researchers to focus solely on designing the \verb|infer()| method to develop new algorithms. Section~\ref{sec:2.3-example-script} will provide a detailed explanation of how to utilize the developed algorithm with just five lines of code.

\subsection{Example Script}
\label{sec:2.3-example-script}
RAGLAB provides a user-friendly interface, allowing users to reproduce RAG algorithms for interaction or evaluation with just five lines of code. In Figure~\ref{fig:raglab-code-demo}, we present an example script for reproducing the Self-RAG algorithm in both interaction and evaluation modes. 
The implementation process is as follows: (1) The \verb|get_config()| function reads parameters from a YAML file and defines the args object. (2) The \verb|SelfRag_Reproduction| class is defined to prepare all settings for the Self-RAG algorithm, based on the args parameters. (3) The \verb|inference()| method in line 9 is called for the interaction mode. (4)
The \verb|inference()| method in line 12 is called again for the evaluation mode.

\section{Experiment}
\label{sec:3-Experiment}
One main aim of RAGLAB is to facilitate fair comparisons among various advanced RAG algorithms. To this end, we conducted comprehensive experiments by employing three distinct base models as generators while maintaining consistency across other fundamental components.

\begin{table*}[ht]
\centering
\caption{Performance comparison of various RAG algorithms using  Llama3-70B as base language model.}
\label{tab:exp-2}
\scriptsize
\setlength{\tabcolsep}{3pt}
\renewcommand{\arraystretch}{1.1}
\begin{tabular*}{\textwidth}{@{\extracolsep{\fill}}lcccccccccccccccccc@{}}
\hline
\multirow{2}{*}{Method} & \multicolumn{2}{c}{PopQA} & \multicolumn{2}{c}{TriviaQA} & \multicolumn{2}{c}{HopPopQA} & \multicolumn{2}{c}{WikiMultiHop} & \multicolumn{2}{c}{ARC} & \multicolumn{2}{c}{MMLU} & \multicolumn{2}{c}{PubHealth} & \multicolumn{2}{c}{StrategyQA} & Factscore & ASQA \\
 & ACC & F1 & ACC & F1 & ACC & F1 & ACC & F1 & ACC & F1 & ACC & F1 & ACC & F1 & ACC & F1 & factscore & str-em \\
\hline
Direct & 25.6& 24.7& 76.4 & 75.6 & 27.8& 38.3 & 28.2 & 34.8 & 90.4 & 68.8 & 73.4& 55.6 & 77.2& 77.2& 70.6& 70.6& 70.5 & 31.99\\
NaiveRag& 39.6& 39& 73.6& 74.2 & 33.8& 44& 28.2 & 38.6 & 89.4 & 67.2 & 70.6& 52.8 & 75.2& 75.2& 63.6& 64 & 84.8 & 27.6\\
RRR & 39& 39.4& 72.8& 73.9 & 31.4& 41.1 & 27.6 & 38.6 & 88.4 & 66.6 & 72.6& 54.6 & 74.4& 74.4& 62.6& 64 & 85.2 & 26.91\\
ITER-RETGEN& 36.2& 40.6& 74.4& 75.4 & 33.6& 46.5 & 26.4 & 35.9 & 89.4 & 67.8 & 72.4& 54.2 & 62.6& 62.6& 59.2& 59.4& 83.9 & 25.32\\
Active Rag & 37& 40& 73.6& 74.7 & 33.2& 43.6 & 26.6 & 36.7 & 89.2 & 67.4 & 71.8& 54.6 & 58& 58& 61 & 61 & 83.7 & 25.96\\
Self Ask& 20.8& 23.6& 65.8& 66.6 & 33.4& 42.9 & 35& 37& 80.4 & 57.4 & 67.4& 48.5 & 60.4& 59.1& 49.6& 55.1& 73.6 & 24.24\\
\hline
\multicolumn{19}{l}{Self-RAG} \\
\renewcommand{\arraystretch}{1.1}
always retrieval& 45.2& 16.8& 77.6 & 43.4 & 40.6 & 26& 38& 22.8  & 89.4& 68 & 72.8& 55.6& 79.4 & 79.4 & 68 & 71.4& 84 & 45.96  \\
adaptive retrieval & 48.8& 17.1& 77.4  & 43.1 & 40.6 & 26& 38.2 & 22.5 & 90 & 68.4& 72.4& 55 & 79.4 & 79.4 & 68 & 71.2& 77.1 & 39.84  \\
no retrieval& 30 & 11.6& 76.6 & 31.9 & 30.8 & 15.5 & 31& 17.2 & 90 & 68.4& 72.6& 55 & 80.4 & 80.4 & 69.4& 69.8& 65.0 & 29.96 \\
\hline
\end{tabular*}
\end{table*}

\begin{table*}[ht]
\centering
\caption{Performance comparison of various RAG algorithms using  GPT3.5 as base language model.}
\label{tab:exp-3}
\scriptsize
\setlength{\tabcolsep}{3pt}
\renewcommand{\arraystretch}{1.1}
\begin{tabular*}{\textwidth}{@{\extracolsep{\fill}}lcccccccccccccccccc@{}}
\hline
\multirow{2}{*}{Method} & \multicolumn{2}{c}{PopQA} & \multicolumn{2}{c}{TriviaQA} & \multicolumn{2}{c}{HopPopQA} & \multicolumn{2}{c}{WikiMultiHop} & \multicolumn{2}{c}{ARC} & \multicolumn{2}{c}{MMLU} & \multicolumn{2}{c}{PubHealth} & \multicolumn{2}{c}{StrategyQA} & Factscore & ASQA \\
 & ACC & F1 & ACC & F1 & ACC & F1 & ACC & F1 & ACC & F1 & ACC & F1 & ACC & F1 & ACC & F1 & factscore & str-em \\
\hline
Direct & 26.6& 13.22& 77& 52.86 & 33.8 & 24.04 & 38& 21.31& 79.6& 21.15  & 63.6& 17.49& 78& 78& 68.2& 68.2& 79.3 & 37.5\\
NaiveRag& 45  & 17.16& 72.8 & 26.47 & 41.6 & 17.74 & 33.2 & 16.44& 67.4& 15.94  & 54.4& 10.83& 53.8  & 53.8  & 61.8& 61.8& 84.5 & 39.1\\
RRR & 46.2& 17.71& 73.6 & 27.45 & 37.2 & 16.34 & 33& 16.43& 68.4& 16.02  & 54.4& 11.01& 54.8  & 54.8  & 63.6& 63.6& 84.5 & 39 \\
ITER-RETGEN & 44.2& 16.75& 73& 26.02 & 44.8 & 18.92 & 34.6 & 16.31& 69.8& 16.95  & 55 & 11.32& 39.2  & 39.2 & 56.2& 56.2& 84.2 & 39.6\\
Active Rag  & 44.2& 17.34& 72.8 & 27.45 & 43.8 & 18.76 & 34.2 & 16.94& 70 & 21.57  & 55.8& 13.38& 50& 50& 61.2& 61.2& 83.7 & 33.7\\
Self Ask& 38.2& 18.89& 68.6 & 38.12 & 36.4 & 25.52 & 41.6 & 23.52& 63.4& 14.71  & 48.6& 10.16& 45.2 & 45.01 & 39.6& 39.43  & 86.7 & 30.2 \\
\hline
\end{tabular*}
\end{table*}

\refstepcounter{experimentcounter} 
\textbf{Experimental \theexperimentcounter{} Generator}:\label{sec:Experiment 1} In Experiment~\ref{sec:Experiment 1} , we select Llama3-8B as the base language model. 
We utilize the open-source data provided by Self-RAG as training data.
 The resulting fine-tuned model is designated as selfrag-llama3-8B, which serves as the generator for the Self-RAG algorithm. To ensure a fair comparison, we removed all special tokens from the training data, then full-weighted fine-tuned another model named Llama3-8B-baseline as the generator for other algorithms. For detailed training parameters, please refer to Appendix~\ref{sec:appendix-a}.

\refstepcounter{experimentcounter} 
\textbf{Experimental \theexperimentcounter{} Generator}:\label{sec:Experiment 2}
In Experiment~\ref{sec:Experiment 2}, we selected Llama3-70B as the base language model. We select the QLoRA\citep{qlora-2023} method to fine-tune selfrag-llama3-70B and Llama3-70B-baseline. We use the same training data as in Experiment~\ref{sec:Experiment 1}. For detailed training parameters, please refer to Appendix~\ref{sec:appendix-c}.

\refstepcounter{experimentcounter} 
\textbf{Experimental \theexperimentcounter{} Generator}:\label{sec:Experiment 3}
In Experiment~\ref{sec:Experiment 3}, we selected GPT3.5 as base model. Additionally, we excluded the Self-RAG algorithm. Because closed-source models are not allowed to add special tokens during the training phase.

\textbf{Additional Experimental Settings}:
We employ ColBERT as the retriever, utilizing Wikipedia 2018 as the external knowledge database. Local models are loaded with float16 precision, and during inference, we fix the random seed and use greedy sampling. The number of retrieved passages and the maximum generation length vary for each benchmark, please refer to Appendix~\ref{sec:appendix-b}.
We strive to maintain consistent instructions across all algorithms. For specific instructions and parameter settings, please refer to Appendix~\ref{sec:appendix-e} and ~\ref{sec:appendix-d}, respectively.
We select 10 comprehensive benchmarks for evaluation experiments, as detailed in Table~\ref{tab:benchmark-datasets}.
Due to limited computation resources, we sequentially sampled 500 data points from each dataset. For evaluation, we employ a range of metrics, including Factscore, ACLE, accuracy (ACC), and F1 score across different datasets.
The task-specific instructions for each dataset are detailed in Appendix~\ref{sec:appendix-f}. 
The results of Experiments~\ref{sec:Experiment 1}, \ref{sec:Experiment 2}, and \ref{sec:Experiment 3} are presented in Tables~\ref{tab:exp-1}, \ref{tab:exp-2}, and \ref{tab:exp-3}, respectively.

\section{Experimental Result and Discussion}
\label{sec:4-exp_result}

After analyzing the results from Experiments~\ref{sec:Experiment 1}, Experiments~\ref{sec:Experiment 2}, and Experiments~\ref{sec:Experiment 3}, we find several valuable insights.
When utilizing selfrag-llama3-8B as the generator for the Self-RAG algorithm, its performance across 10 benchmarks did not significantly surpass other RAG algorithms. However, when employing selfrag-llama3-70B as the generator, the Self-RAG algorithm significantly outperformed others in 10 benchmarks.
We also find that Naive RAG, RRR, Iter-RETGEN, and Active RAG demonstrate comparable performance across 10 datasets. Notably, the ITER-RETGEN algorithm exhibits superior performance in Multi-HopQA tasks.
Furthermore, our findings indicate that RAG systems underperform compared to direct LLMs in multiple-choice question tasks. This conclusion aligns with experimental results from other studies\citep{RQ-RAG,selfrag,blendfilter-2024}. 
A possible explanation for this phenomenon is that multiple-choice questions include both the question and candidate answers. Additional retrieved information may mislead the generator.


\section{Human Evaluation of RAGLAB}

To comprehensively evaluate the user experience of the RAGLAB library, we implemented a user study. We developed a questionnaire comprising 12 questions, as shown in Figure~\ref{fig:appendix-G-user_questionnaire}. The study participants consisted of 20 NLP researchers, each having utilized RAGLAB at least three days. The questionnaire was administered offline, achieving a \verb|100%| response rate. The results indicated that \verb|85%| of respondents perceived RAGLAB as significantly enhancing their research efficiency, and \verb|90%| expressed willingness to recommend RABLAB to other researchers. Additionally, we gathered valuable suggestions for improvement, which will guide future system development.

\section{Conclusion}

We introduced RAGLAB, an efficient and user-friendly library for the fair comparison of RAG algorithms.  With RAGLAB, researchers can easily conduct fair comparisons of existing RAG algorithms and develop new algorithms. We also conducted a fair comparison of 6 widely used RAG algorithms across 10 benchmarks, finding several valuable insights. We believe RAGLAB will become an essential research tool for the NLP community.

\section*{Limitations}
RAGLAB provides a comprehensive and fair comparison of various RAG algorithms performance. However, there remain potential improvement in future work. 

Due to limited computational resources, RAGLAB currently encompasses only 6 algorithms and 10 widely used benchmarks. However, there remains a need to include more algorithms and datasets. In future work, we will continue to follow the latest research developments and incorporate novel algorithms for fair comparison.

During the implementation process, we found that different retriever models and external knowledge databases significantly impact the performance of RAG algorithms. Due to limited computational resources, we only processed Wikipedia 2018 and Wikipedia 2023. In future work, we plan to include a wider variety of knowledge databases and conduct experiments on how different retriever models and external knowledge databases influence the performance of RAG algorithms.

Current research primarily focuses on improving the performance of algorithms, lacking a comprehensive evaluation of resource consumption and inference latency. At present, RAGLAB incorporates only 3 classic metrics and 2 advanced metrics. In future work, we aim to expand our evaluation framework to include a more diverse range of metrics.

We encourage the open-source community to address these limitations together. Our objective is to continually refine the RAGLAB framework, aiming to provide the most efficient and reliable evaluation platform and development tools.

\newpage
\bibliography{anthology,custom}
\bibliographystyle{acl_natbib}

\clearpage
\appendix
\section{Training Parameters for Llama3-8B}
\label{sec:appendix-a}

This appendix outlines the key training parameters used for fine-tuning the Llama3-8B model in our experiments. We employed full-weight fine-tuning on the Llama3-8B base model. The maximum sequence length was set to 4096 tokens, with a learning rate of 2e-5 and training conducted for 1 epoch.  For a comprehensive list of training parameters, including computational resources and optimization settings, please refer to Table \ref{tab:4-training_params}.

\begin{table}[h]
\centering
\caption{Training Parameters for Llama3-8B.}
\begin{tabular}{lc}
\hline
\textbf{Parameter} & \textbf{Value} \\
\hline
Model & Llama3-8B \\
Fine-tuning method & Full weight \\
Number of GPUs & 8 \\
Total batch size & 32 \\
Batch size per GPU & 1 \\
Gradient accumulation steps & 4 \\
Mixed precision & bf16 \\
Maximum sequence length & 4096 \\
Learning rate & 2e-5 \\
Learning rate scheduler & Linear \\
Warmup ratio & 0.03 \\
Weight decay & 0 \\
Number of epochs & 1 \\
DeepSpeed ZeRO stage & 3 \\
\hline
\end{tabular}
\label{tab:4-training_params}
\end{table}

\section{Inference Parameters for Different Benchmarks}
\label{sec:appendix-b}

The number of retrieved passages and the maximum generation length were adjusted for each benchmark to accommodate their specific requirements. Table \ref{tab:inference_params} presents a comprehensive overview of these parameters across various benchmarks.

\begin{table}[h]
\centering
\caption{Inference Parameters for Different Benchmarks.}
\small
\begin{tabular}{lccc}
\hline
\textbf{Benchmark} & \textbf{Precision} & \textbf{Max Length} & \textbf{N Docs} \\
\hline
PopQA & float16 & 300 & 10 \\
TriviaQA & float16 & 300 & 10 \\
HotpotQA & float16 & 300 & 10 \\
2WikiMultiHopQA & float16 & 300 & 10 \\
Arc & float16 & 50 & 10 \\
PubHealth & float16 & 50 & 10 \\
MMLU & float16 & 50 & 10 \\
StrategyQA & float16 & 300 & 10 \\
Factscore & float16 & 300 & 5 \\
ASQA & float16 & 300 & 5 \\
\hline
\end{tabular}
\label{tab:inference_params}
\end{table}

\newpage
\section{Training Parameters for Llama3-70B}
\label{sec:appendix-c}

 We employed QLoRA fine-tuning on the Llama3-70B base model with a 4-bit quantization. The maximum sequence length was set to 4096 tokens, with a learning rate of 2e-5 and training conducted for 1 epoch. For the LoRA configuration, we used a rank of 64, an alpha of 16, and a dropout of 0.1. These parameters were consistently applied for both the self-rag-llama3-70B and Llama3-70B-baseline models. The training data remained the same as in Experiment 1. For a comprehensive list of training parameters, please refer to Table \ref{tab:training_params_llama3_70b}.

\begin{table}[ht]
\centering
\caption{Training Parameters for Llama3-70B using QLoRA.}
\footnotesize
\begin{tabular}{lc}
\hline
\textbf{Parameter} & \textbf{Value} \\
\hline
Model & Llama3-70B \\
Fine-tuning method & QLoRA \\
Total batch size & 32 \\
Batch size per GPU & 1 \\
Gradient accumulation steps & 4 \\
Mixed precision & bf16 \\
Maximum sequence length & 4096 \\
Learning rate & 2e-5 \\
Learning rate scheduler & Linear \\
Warmup ratio & 0.03 \\
Weight decay & 0 \\
Number of epochs & 1 \\
Quantization & 4-bit \\
Quantization type & fp4 \\
LoRA rank & 64 \\
LoRA alpha & 16 \\
LoRA dropout & 0.1 \\
\hline
\end{tabular}
\label{tab:training_params_llama3_70b}
\end{table}

\section{Configuration details for RAG methods}
\label{sec:appendix-d}

This appendix provides detailed configuration information for the various Retrieval-Augmented Generation (RAG) methods employed in our experiments. All algorithm parameters were set according to the optimal values reported in their respective original papers to ensure fair comparison and optimal performance.

\begin{table}[htbp]
\centering
\small
\caption{Configuration details for RAG methods.}
\label{tab:rag-config}
\begin{tabular}{lcc}
\hline
Method & Parameter & Value \\
\hline
\multirow{6}{*}{Self-RAG} 
 & beam\_width & 2 \\
 & max\_depth & 7 \\
 & w\_rel & 1.0 \\
 & w\_sup & 1.0 \\
 & w\_use & 0.5 \\
 & threshold & 0.2 \\
\hline
\multirow{3}{*}{Active RAG} & filter\_prob & 0.8 \\
 & masked\_prob & 0.4 \\
 & Query formulation & Implicit \\
\hline
\multirow{1}{*}{ITER-RETGEN} & max\_iteration & 3 \\
\hline
Self Ask & max\_iteration & 5 \\
\hline
\end{tabular}
\end{table}

\newpage
\section{Algorithm Instructions}
\label{sec:appendix-e}

\begin{figure}[!ht]
    \centering
    \includegraphics[scale=0.5]{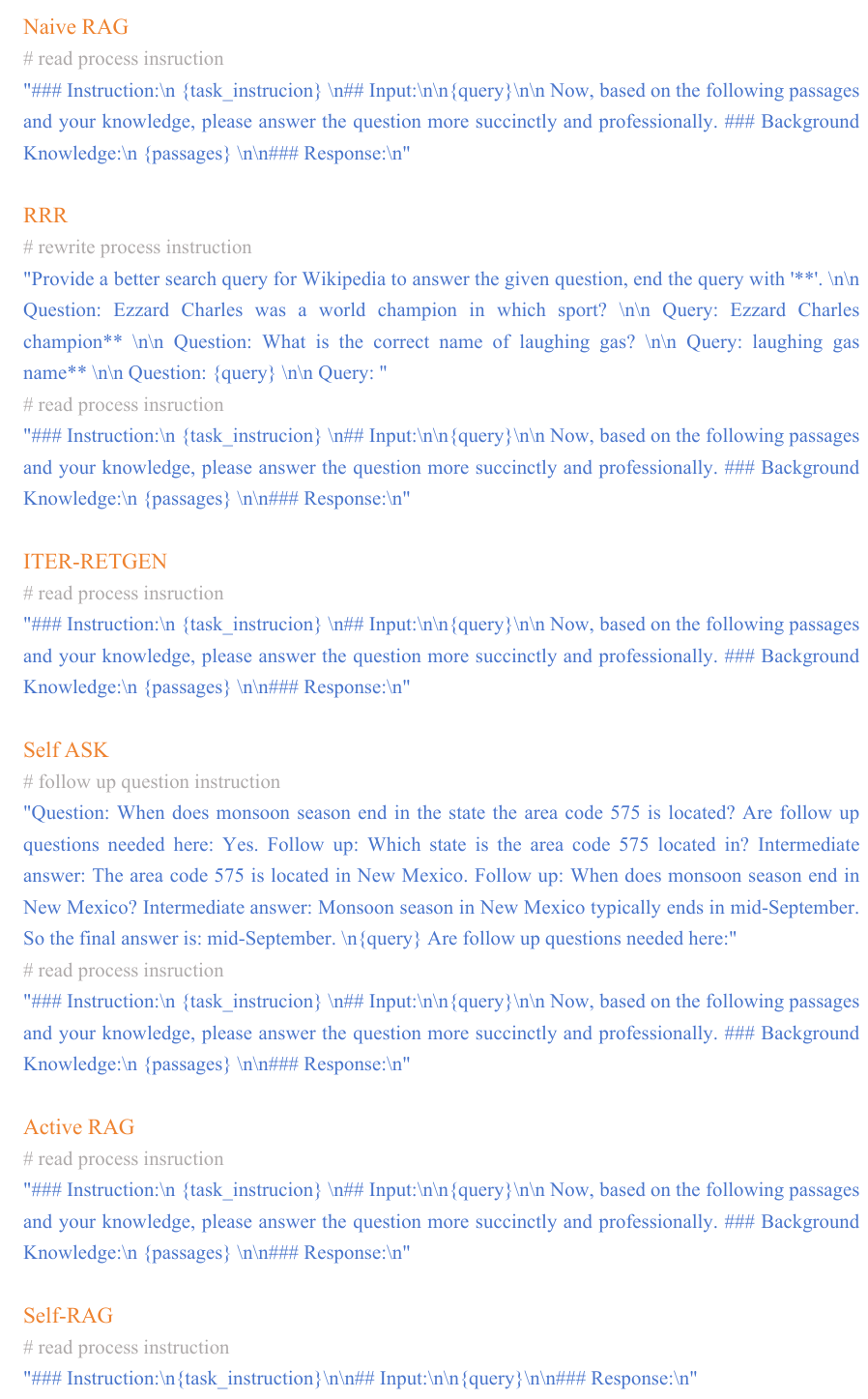}
    \caption{Algorithm Instructions.}
    \label{fig:Appendix-Instructions}
\end{figure}

\section{Datasets Instructions.}
\label{sec:appendix-f}

\begin{figure}[!ht]
    \centering
    \includegraphics[scale=0.42]{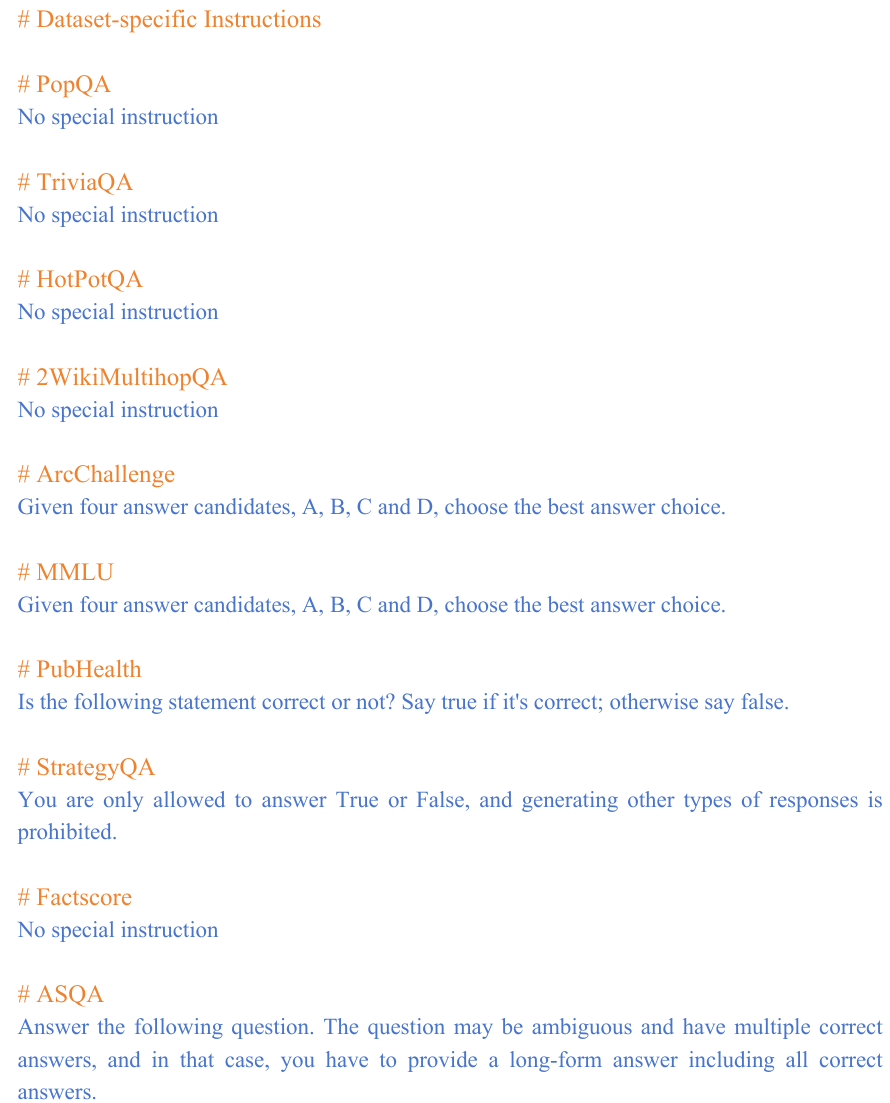}
    \caption{Datasets Instructions.}
    \label{fig:Datasets-Instructions}
\end{figure}

\section{User Evaluation Questionnaire}
\begin{figure}[!ht]
    \centering
    \includegraphics[scale=0.5]{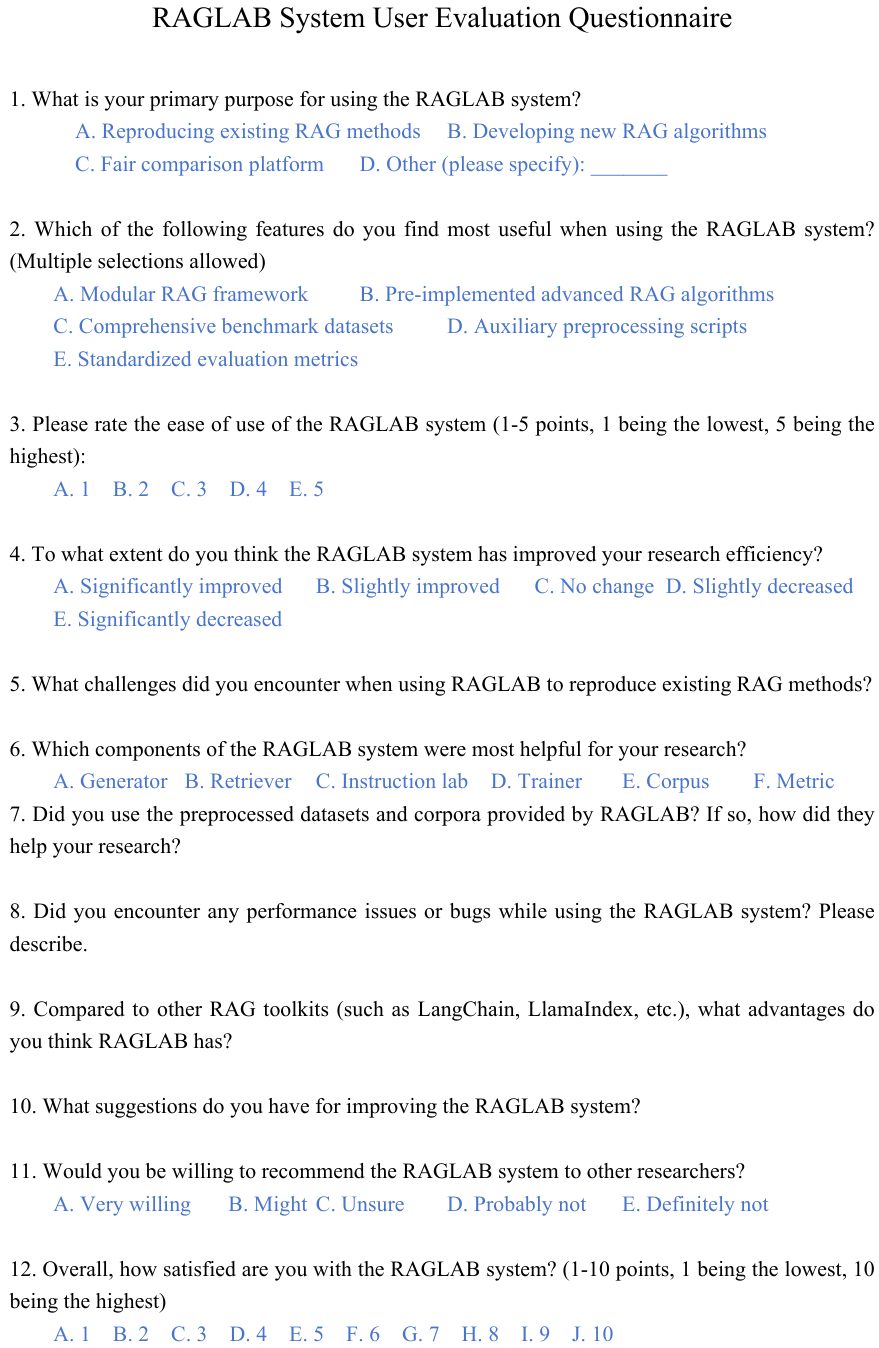}
    \caption{RAGLAB System User Evaluation Questionnaire}
    \label{fig:appendix-G-user_questionnaire}
\end{figure}

\end{document}